%% file: main.tex
\documentclass[twoside]{article}

\usepackage[accepted]{aistats2022}

\usepackage{import}
\import{misc}{preamble.tex}

\usepackage[round]{natbib}

\begin{document}

\twocolumn[

\aistatstitle{Learning from Multiple Noisy Partial Labelers}

\aistatsauthor{ Peilin Yu\And Tiffany Ding \And  Stephen H. Bach}

\aistatsaddress{ Brown University \And  UC Berkeley \And Brown University } ]

\import{sections}{abstract.tex}
\import{sections}{intro}

\import{sections}{related}
\import{sections}{method}

\import{sections}{experiments}

\import{sections}{conclusion}

\import{sections}{acknowledgements}
{\small
\bibliography{cite.bib}
}

\clearpage
\appendix

\thispagestyle{empty}

\onecolumn \makesupplementtitle

\import{sections}{appendix}

\end{document}

%% file: misc/preamble.tex
\usepackage[mathscr]{euscript}

\usepackage{changepage}
\usepackage{graphicx}
\usepackage{amsmath}
\usepackage{bbm}
\usepackage{color}
\usepackage{amssymb}
\usepackage{amsthm}
\usepackage{booktabs}

\usepackage[frozencache=true,cachedir=minted-cache]{minted} 
\usepackage{multicol}
\usepackage{hyperref}

\DeclareMathOperator*{\argmax}{argmax}

\theoremstyle{definition}
\newtheorem{example}{Example}
\newtheorem{theorem}{Theorem}

\renewcommand{\paragraph}[1]{\textbf{#1}\;\; }

%% file: sections/abstract.tex
\begin{abstract}
Programmatic weak supervision creates models without hand-labeled training data by combining the outputs of heuristic labelers.
Existing frameworks make the restrictive assumption that labelers output a single class label.
Enabling users to create partial labelers that output subsets of possible class labels would greatly expand the expressivity of programmatic weak supervision.
We introduce this capability by defining a probabilistic generative model that can estimate the underlying accuracies of multiple noisy partial labelers without ground truth labels.
We show how to scale up learning, for example learning on 100k examples in one minute, a $300\times$ speed up compared to a naive implementation.
We also prove that this class of models is generically identifiable up to label swapping under mild conditions.
We evaluate our framework on three text classification and six object classification tasks.
On text tasks, adding partial labels increases average accuracy by 8.6 percentage points.
On image tasks, we show that partial labels allow us to approach some zero-shot object classification problems with programmatic weak supervision by using class attributes as partial labelers.
On these tasks, our framework has accuracy comparable to recent embedding-based zero-shot learning methods, while using only pre-trained attribute detectors.
\end{abstract}

%% file: sections/intro.tex
\section{INTRODUCTION}
The need for large-scale labeled datasets has driven recent research on methods for \emph{programmatic weak supervision} (PWS), such as data programming~\citep{ratner:neurips16, ratner:vldbj20}, adversarial label learning~\citep{arachie:jmlr21}, learning rules from labeled exemplars~\citep{awasthi:iclr20}, and weak supervision with self-training~\citep{karamanolakis:naacl21}.
In PWS, \emph{labeling functions}, such as user-written rules and other heuristics, provide votes on the true labels for unlabeled examples or abstain from voting.
Then, a label model, such as a probabilistic generative model or minimax game, is often used to estimate the true labels in a way that accounts for unknown differences in accuracies and other properties of the labeling functions.
Finally, these estimated labels are used to train an end model on the unlabeled data to generalize beyond the information contained in the labeling functions.
This approach has had recent success with applications in natural language processing~\citep{mallinar:aaai19, safranchik:aaai20}, computer vision~\citep{chen:iccv19}, medicine~\citep{fries:natcomm19, saab:npjdigmed20, fries:natcomm21}, and the Web~\citep{bach:sigmod19-industrial}.
However, all of these methods assume that labeling functions cast votes for \emph{individual} classes. 
In this work, we propose to generalize PWS to support labeling functions that cast votes for a \emph{subset} of classes, called partial labels.
We refer to such labeling functions as \emph{partial labeling functions} (PLFs).
Our goal is to aggregate information from multiple partial labeling functions that are noisy (i.e., have imperfect accuracy) in order to estimate labels for unlabeled data.

\begin{figure}[t]
  \centering
  \includegraphics[width=0.475\textwidth]{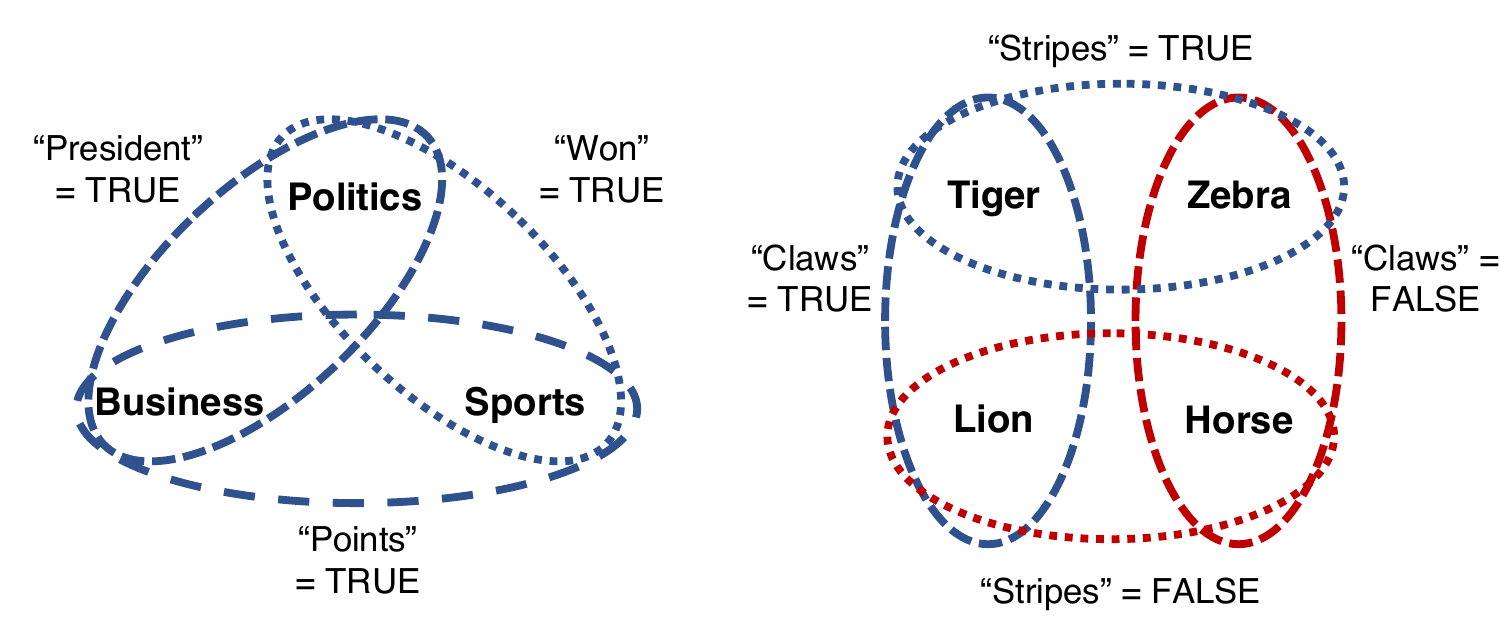}
  \caption{Examples of the expressivity of partial labeling functions. On the left, three functions each vote for two of three classes if they observe a particular token in a news article and abstain otherwise. On the right, two functions each vote for two of four classes if they detect a particular attribute in an image. Otherwise, they vote for the other two.}
  \label{fig:example}
\end{figure}

Incorporating partial labels into PWS would enable users to take advantage of a wider range of domain knowledge.
In typical PWS frameworks, only heuristics that are specific to one class can be incorporated.
As a result, creating labeling functions requires careful task-specific engineering to avoid features that are shared by more than one class.
For example, consider the task of classifying images of animals with a label from the set \{\texttt{HORSE, TIGER,  LION, ZEBRA}\}.
There are many useful heuristics that can be learned from other labeled data sets, such as detectors for claws or stripes~\citep{lampert:cvpr09}.
Such heuristics divide the label space with multiple partitions.
A claw detector could produce two partial labels: \{\texttt{TIGER, LION}\} if a claw is detected and \{\texttt{HORSE, ZEBRA}\} if not.
Likewise, a stripe detector could output \{\texttt{TIGER, ZEBRA}\} if stripes are detected and \{\texttt{HORSE, LION}\} if not (Figure~\ref{fig:example}).
However, these heuristics cannot be used as labeling functions in current PWS frameworks.
More generally, we observe a need for partial labeling functions in many multiclass applications where users want to express heuristics that narrow down the set of possible class labels but are not specific to a single class.

Learning from multiple noisy PLFs is challenging because we must resolve ambiguity arising from three sources: (1) PLF imprecision, i.e., voting for a set of classes instead of a single class, (2) PLF inaccuracy, i.e., voting for a set of classes that does not contain the true class, and (3) conflict among multiple PLFs.
A further requirement is that PWS frameworks should support labeling functions that abstain, meaning they can choose not to label certain examples.
This is particularly critical for hand-engineered rules that might be highly specialized.
A framework for learning from multiple noisy PLFs should therefore be able to resolve all these types of ambiguities in a principled way while also maintaining the expressive capabilities of existing PWS frameworks.

This problem setting is quite general and is related to multiple lines of work in machine learning, although each of them only addresses part of the problem considered here.
As mentioned above, previous PWS frameworks generally require labeling functions to provide a single class label~\citep{ratner:neurips16,ratner:vldbj20,arachie:jmlr21,awasthi:iclr20,karamanolakis:naacl21,safranchik:aaai20}.
One exception is Snorkel MeTaL~\citep{ratner:aaai19}, which is capable of handling labeling functions with a multi-task tree structure where higher-level labeling functions are grouped into super classes that encompass fine-grained classes.
This requirement of a tree structure makes modeling partial labels that divide the label space into overlapping subsets practically infeasible.
There is also a wide body of work on learning from partial labels, also called superset learning~\citep{jin:neurips02,nguyen:kdd08,luo:neurips10,cour:jmlr11,liu:neurips12,liu:icml14,hullermeier:ecml15,cabannnes:icml20,wang:ijcai20,cabannes:arxiv21,zhang:aaai21}.
In these settings, there is generally one partial label per example.
The ambiguity in the imprecise labels in such settings can be resolved as a maximum likelihood or risk minimization problem.
Many methods additionally learn the likely confusions between partial and true labels~\citep{durand:cvpr19,li:cikm20,yan:aaai20,xie:pami21}.
However, such methods do not handle the case of multiple partial labelers that can disagree and abstain.
Finally, some work on zero-shot learning (ZSL) creates attribute detectors that can be viewed as partial labelers~\citep{farhadi:cvpr09, lampert:cvpr09, palatucci:neurips09,jayaraman:neurips14}.
PWS with partial labels can also be viewed as a generalization of the transductive ZSL setting~\citep{xian:pami18,wang:tist19}, in which labelers are allowed to abstain and a class may be associated with multiple attribute values.
Across all these areas, there remains a need for learning from multiple noisy partial labelers.

To address this issue, we propose a generalized PWS framework that supports partial labels and handles the additional ambiguity caused by the imprecise outputs of the PLFs.
First, we introduce a probabilistic generative model that estimates the agreement between the outputs of each partial labeling function and the latent, true label.
Second, we show how to learn the model parameters efficiently for large datasets. 
For example, we can learn on 100k examples with 10 PLFs in one minute.
Since PWS is inherently human-in-the-loop, fast iteration is crucial.
Third, we prove that this model's parameters are generically identifiable up to label swapping under mild conditions on the PLFs.
This result means that we can estimate the accuracy of each partial labeling function without access to ground truth labels in a principled way.
Using the learned parameters, we can compute the posterior distribution over true labels for each example.
These probabilistic training labels can then be used to train an end model in the same manner as other PWS frameworks.

We demonstrate this framework with experiments on three text and six object classification tasks.
On the text classification tasks, we show that the additional flexibility provided by partial labelers enables heuristics that significantly improve over single-class labelers alone.
We find an average 8.6 percentage point improvement in accuracy.
On the object classification tasks, we find that modeling the accuracies of the PLFs explicitly enables us to achieve accuracy comparable to recent embedding-based ZSL methods using only pre-trained attribute detectors.
These results provide a foundation for constructing and learning more modular, reusable knowledge sources for weak supervision.

%% file: sections/related.tex
\section{RELATED WORK}

In the past few years, programmatic weak supervision (PWS) has emerged as a systematic approach to efficiently create labeled training data \citep{ratner:neurips16, ratner:vldbj20, arachie:jmlr21, awasthi:iclr20, karamanolakis:naacl21}.
A typical PWS framework consists of three stages.
First, domain experts engineer weak supervision sources in the form of labeling functions, such as rules or classifiers related to the target task.
Second, a label model, such as a probabilistic generative model, is used to estimate the latent true labels using the labeling function outputs.
Third, the estimated labels are used to train an end model that generalizes beyond the information in the supervision sources.
The core of a typical PWS framework is the label modeling stage.
The choice of label model determines what types of supervision sources are supported.
Many frameworks are based on crowdsourcing methods~\citep{dawid:royalstats79, nitzan:inteconreview82,gao:arxiv13}, where providing a single label is a natural assumption.
In the original data programming framework~\citep{ratner:neurips16}, labeling functions can output a single label or abstain.
The label model is generative, meaning that each true label is a latent variable and the observed votes of the labeling functions are conditioned on the true labels.
The parameters of the label model are learned by maximizing the marginal likelihood of the observed votes.
Statistical dependencies such as correlations among the votes can be modeled, and methods exist to learn specific types of dependencies from unlabeled data~\citep{bach:icml17, varma:icml19}.

The Snorkel MeTaL~\citep{ratner:aaai19} framework extends data programming to learn across multiple, related tasks organized in a tree structure.
For example, in a fine-grained named entity recognition task, one might use a set of labeling functions that vote on coarse-grained entity types and separate sets of labeling functions to further vote on the subtypes within each coarse type.
The outputs of labeling functions at higher levels of the tree can be thought of as a restricted form of partial labels, in the sense that all labeling functions must follow the same tree-structured organization of the classes.
In contrast, in our setting, each partial labeling function can organize the classes into its own, possibly overlapping groups.

Other PWS frameworks have approached labeling functions and the label modeling processes in different ways, but all so far assume that each labeling function votes for a single class.
Adversarial label learning~\citep{arachie:jmlr21}, performance-guaranteed majority vote~\citep{mazzetto:aistats21}, adversarial multi class learning~\citep{mazzetto:icml21}, and related work in semi-supervised ensemble learning~\citep{balsubramani:colt15, balsubramani:neurips15, balsubramani:neurips16} solve minimax games based on assumed or estimated constraints on labeling function accuracies.
\citet{awasthi:iclr20} proposed learning from rules and exemplars for those rules, learning to downweight the confidence in the rules on data instances not similar to the exemplars.
\citet{karamanolakis:naacl21} proposed integrating PWS with semi-supervised self-training.

Other work on learning with partial labels has focused on the case where there is a single partial label per example.
Classifiers can be learned via maximum likelihood estimation~\citep{jin:neurips02,liu:neurips12} or empirical risk minimization~\citep{nguyen:kdd08,luo:neurips10,cour:jmlr11,liu:icml14,hullermeier:ecml15,cabannnes:icml20,cabannes:arxiv21,feng:neurips2020}.
Many methods additionally learn the likely confusions between partial and true labels~\citep{durand:cvpr19, li:cikm20, yan:aaai20, xie:pami21}.
\citet{wang:ijcai20} proposed learning multiple partially labeled tasks simultaneously, in order to exploit structure among the tasks, but during training there is still only one partial label per prediction.
Partial labels are also related to complementary labels~\citep{ishida:neurips17, feng:icml20}, which are annotations that indicate which label the example does \emph{not} have.

Our problem setting is also related to some forms of zero-shot learning (ZSL)~\citep{xian:pami18,wang:tist19}.
In zero-shot classification, a model learns to match semantic descriptions of classes to examples of those classes.
Once learned, the model can be applied to novel classes.
Many early approaches to ZSL created detectors for different attributes~\citep{farhadi:cvpr09, lampert:cvpr09, palatucci:neurips09,jayaraman:neurips14}.
In the transductive setting~\citep{xian:pami18,wang:tist19}, in which the target classes are known and unlabeled examples of them are available during model development, these detectors can be viewed as restricted partial labeling functions that always divide the label set into non-overlapping groups and never abstain.
More recently, much work on ZSL has moved away from relying entirely on attribute detectors, and recent work can be grouped into either embedding-based or generative-based methods~\citep{pourpanah:survey2020}.
Embedding-based methods align representation spaces between classes and examples in order to classify unlabeled data~\citep{socher:neurips13, frome:neurips13, romera:icml15,xian:cvpr16, xian:pami18,wang:tist19}.
Some work, e.g., \citet{liu:iccv19} and \citet{liu:aaai20}, also learn to exploit and expand attribute-based information, but generally still do not use separate attribute detectors.
On the other hand, generative-based ZSL methods generate examples of the unseen classes with deep generative models and then train a classifier with that data~\citep{bucher:cvpr17, verma:cvpr18, felix:eccv18, sariyildiz:cvpr19,xian:cvpr19,narayan:eccv20}.
In our experiments, we compare with transductive embedding-based methods, which are more similar to PWS because both involve trying to label a fixed, unlabeled data set.
We leave incorporating zero-shot data generation into PWS for future work.

%% file: sections/method.tex
\section{A FRAMEWORK FOR PLFs}

Following prior work in PWS~\citep{ratner:neurips16,ratner:vldbj20}, our framework consists of three stages.
First, we define partial labeling functions as sources of weak supervision (Section~\ref{section:plf_dev}).
Second, we propose and analyze a label model to aggregate their outputs (Section~\ref{section:label_model}).
Third, the learned label model is used to compute the posterior distribution for the true label of each unlabeled example, which is used to train a noise-aware classifier (Section~\ref{section:end_model}).

\subsection{Partial Labeling Functions}
\label{section:plf_dev}

We propose generalizing labeling functions to \emph{partial labeling functions} (PLF) in order to make use of many available weak supervision sources that are informative but not specific enough to identify a single class.
PLFs can range in granularity, from dividing the label space into two large groups down to identifying a specific class, i.e., a regular labeling function.
This flexibility allows users to take advantage of many additional supervision signals, as we illustrate in our experiments.

A PLF $G$ is a function that maps an unlabeled example to a proper subset of the possible labels or abstains by outputting the full set of all possible labels.
Formally, our goal is to learn a classifier $C: \mathcal{X} \rightarrow \mathcal{Y}$, where $\mathcal{X}$ is the space of inputs and $\mathcal{Y} = \{y_1, \dots, y_k\}$ is the set of possible labels.
A PLF is then a function $G: \mathcal{X} \rightarrow \mathcal{G} \subseteq \mathbb{P}(\mathcal{Y}) \setminus \{\emptyset\}$, where $\mathbb{P}(\mathcal{Y})$ is the power set of $\mathcal{Y}$.
$G(X)$ is a partial label for $X \in \mathcal{X}$, i.e., the set of labels that the PLF indicates the example $X$ could have (although this information could be incorrect).
If $G(X) = \mathcal{Y}$, the PLF is said to abstain, because it provides no information about the true label.
As described further in Section~\ref{section:label_model}, a key characteristic of a PLF is its codomain $\mathcal{G}$, excluding when the PLF abstains.
We denote this set of partial labels for a PLF $G$ as $T(G)$ = $\mathcal{G} \setminus \{\mathcal{Y}\}$.
To ensure that our label model is well-defined, we will impose these conditions on $T(G)$: (1) each label $y \in \mathcal{Y}$ appears in at least one element of $T(G)$, and (2) no label $y \in \mathcal{Y}$ appears in every element of $T(G)$.
These are very mild conditions that can easily be satisfied by adding a ``dummy'' output to the codomain $\mathcal{G}$ that the PLF might not actually produce.
A PLF can be defined based on a variety of noisy supervision heuristics using domain knowledge and/or available resources, such as classifiers for related tasks.
To better understand PLFs, consider the following text and object classification examples corresponding to Figure~\ref{fig:example}: 
\begin{example}
Consider a news classification task where $\mathcal{Y} = \texttt{\{POLITICS, SPORTS, BUSINESS\}}$.
In this task, some words can be very informative as supervision sources even if they do not narrow the example down to a specific class.
For example, the word ``president'' may frequently appear in both political and business contexts.
We can construct a PLF $G$ based on a simple token matcher for ``president'' such that $G: \mathcal{X} \rightarrow \{ \{\texttt{BUSINESS}, \texttt{POLITICS}\}, \{\texttt{SPORTS}\}, \mathcal{Y} \}$.
If the token ``president'' appears in the example $X$, then $G(X) = $\{\texttt{BUSINESS}, \texttt{POLITICS}\}.
Otherwise, $G(X) = \mathcal{Y}$, i.e., $G$ abstains, because the absence of the token is not enough to conclude anything about the label with high confidence.
In this example, $T(G) = \{ \{\texttt{BUSINESS}, \texttt{POLITICS}\}, \{\texttt{SPORTS}\}\}$.
Notice here \{\texttt{SPORTS}\} is a ``dummy'' label set to satisfy the conditions on $T(G)$ described above.
\end{example}

\begin{example}
Consider an object classification task where $\mathcal{Y} = \{ \texttt{HORSE}, \texttt{TIGER}, \texttt{LION}, \texttt{ZEBRA} \}$.
Following work in zero-shot learning, we can build a binary classifier for the visual attribute of having stripes by training on other classes of animals for which we already have labels.
We can then use the classifier's output to define a PLF 
$G_1: \mathcal{X} \rightarrow \{ \{ \texttt{TIGER}, \texttt{ZEBRA} \}, \{ \texttt{HORSE}, \texttt{LION} \} \}$.
For an example $X$, if the stripes detector returns a positive label, then $G_1(X) = \{ \texttt{TIGER}, \texttt{ZEBRA} \}$.
Otherwise, $G_1(X) = \{ \texttt{HORSE}, \texttt{LION} \}$.
We can similarly construct a PLF with a claw detector as $G_2: \mathcal{X} \rightarrow \{ \{ \texttt{TIGER}, \texttt{LION} \}, \{ \texttt{HORSE}, \texttt{ZEBRA} \} \}$.
\end{example}

PLFs are a generalization of the labeling functions used in prior work on PWS; traditional labeling functions can be represented as PLFs with codomain $\mathcal{G} = \{ \{y_1\}, \dots, \{y_k\}, \mathcal{Y} \}$.
PLFs provide the additional flexibility to users of incorporating weak supervision heuristics with differing granularitities and ways of dividing the label space.

\subsection{Label Model}
\label{section:label_model}
In our framework, users provide two inputs: PLFs and unlabeled examples in $\mathcal{X}$.
Like other PWS frameworks, at the core of our method is a probabilistic label model that captures the properties of the weak supervision sources by representing the unknown ground-truth labels as latent variables.
In this subsection, we propose and analyze a probabilistic label model for PLFs.
Our label model is straightforward to define as a generalization of existing ones for labelers that provide a single label~\citep{dawid:royalstats79, ratner:neurips16}.
It defines a correct output of a partial labeler as one that is consistent with the unknown, true label, i.e., that the true label is in the output set.
If the partial labeler is trivially correct because it outputs the set of all possible labels, it is said to abstain, which is not counted towards its estimated accuracy.

While straightforward to define, using the model introduces new challenges.
The first challenge is practical.
Existing methods for optimizing the marginal likelihood of the label model in the single-label case do not work for PLFs.
The matrix factorization approach proposed by~\citet{ratner:aaai19} exploits the fact that the matrix of agreements among labelers can be expressed as a product of the model parameters, but this does not hold in the PLF case.
\citet{bach:sigmod19-industrial} proposed optimizing the likelihood in an auto-differentiation framework like PyTorch~\citep{paszke:neurips-ws17}.
We find that naively applying this approach is impractically slow, and that carefully defining the computation graph leads to a $300\times$ speedup.

The other challenge we address is theoretical.
We consider when it is possible to learn the parameters of the model without access to ground truth in a principled way, i.e., when the model is identifiable.
Prior theoretical work either on PWS or learning with partial labels is not applicable to our scenario.
Prior work on identifiability for PWS does not consider partial labels~\citep{ratner:aaai19,safranchik:aaai20}, and new conditions and arguments are needed to handle them.
Prior work on risk bounds for learning with partial labels does not address noise nor multiple labelers~\citep{cour:jmlr11,liu:icml14,hullermeier:ecml15,feng:neurips20}.
Therefore, our main contributions in this section are a fast learning technique and a theorem characterizing sufficient identifiability conditions for our model.

\paragraph{Setup}
For a classification task with input space $\mathcal{X}$ and label space $\mathcal{Y} = \{y_1, \dots, y_k\}$, we are given $m$ unlabeled examples $X = (X_1, \dots, X_m)$ with unknown ground truth labels $Y = (Y_1,\dots,Y_m)$ such that $(X,Y)$ are i.i.d.~samples from some distribution $\mathcal{D}$.
We are also given $n$ PLFs $G = (G_1, \dots, G_n)$.
We use $G$ as shorthand for the $m \times n$ array of PLF outputs where $G_{ai} = G_i(X_a)$ when it is clear from context.

\paragraph{Joint Distribution}
We define a joint distribution $P(G, Y)$ over the outputs of the PLFs on $X$ and the latent, true labels $Y$.
Like prior work~\citep{ratner:neurips16}, we assume that the PLF outputs are conditionally independent given the true labels, i.e., the naive Bayes assumption.
In practice this works well, but extending work on learning more complex distributions for other types of PWS is a potential direction for future exploration~\citep{bach:icml17,varma:icml19}.
Analogous to prior work~\citep{ratner:aaai19}, for each PLF $G_i$, we define parameters $\alpha_i \in [0,1]^k$ and $\beta_i \in [0,1]$.
Each element $\alpha_{ij}$ is the accuracy of $G_i$ on examples of class $y_j$, i.e., the probability that $y_j \in G_{ai}$ given that $X_a$ has label $y_j$ and $G_{ai}$ is not $\mathcal{Y}$.
$\beta_i$ is the propensity of $G_i$ voting, i.e., not abstaining.
In other words, $\beta_i = P(G_{ai} \neq \mathcal{Y})$.
In our framework, the class balance $P(Y)$ can either be a learned distribution or fixed.
We assume that if a PLF $G_i$ makes a mistake, it outputs an incorrect partial label from $T(G_i)$ uniformly at random.

To define the joint distribution $P(G, Y)$, for each PLF $G_i$ we also need to refer to the sets in $T(G_i)$ that are consistent or inconsistent with each label.
Let $N_{ij} = \{\mathcal{L} \vert y_j \in \mathcal{L} \text{\ for\ } \mathcal{L} \in T(G_i) \}$ be the set of label sets in the codomain of $G_i$ that contain label $y_j$ (excluding $\mathcal{Y}$).
Likewise, let $N_{ij}^\mathrm{C} = \{\mathcal{L} \vert  y_j \notin \mathcal{L} \text{\ for\ } \mathcal{L} \in T(G_i) \}$ be the set of label sets in the codomain of $G_i$ that do not contain label $y_j$.
Then, the joint distribution is
\begin{equation}
\label{eq:joint}
P(G, Y) =
    \displaystyle\prod_{a=1}^{m}
    P(Y_a) \displaystyle\prod_{i=1}^{n} P(G_{ai} | Y_a)
\end{equation}
where
\begin{equation}
\label{eq:conditional}
    P(G_{ai}|Y_a = y_j) = 
    \begin{cases}
      1 - \beta_i, & \text{if}\ G_{ai} = \mathcal{Y} \\[.75em]
      \frac{\beta_i\alpha_{ij}}{|N_{ij}|}, & \text{if}\ y_j \in G_{ai} \neq \mathcal{Y} \\[1em]
      \frac{\beta_i(1-\alpha_{ij})}{|N_{ij}^\mathrm{C}|}, & \text{if}\ y_j \notin G_{ai}.
    \end{cases}
\end{equation}

\paragraph{Learning}
Given the unlabeled examples and PLF outputs G, our goal is to estimate the parameters of $P(G, Y)$ (denoted collectively as $\Theta$) and compute the posterior $P(Y|G)$ over the unknown labels.
To estimate $\Theta$, we maximize the marginal likelihood of the observed outputs of the PLFs:
\begin{equation}
\hat{\Theta} = \argmax_{\Theta} P_{\Theta} (G) = \argmax_{\Theta} \displaystyle\sum_{Y} P_{\Theta} (G, Y) \; .
\end{equation}
This optimization is implemented in PyTorch~\citep{paszke:neurips-ws17}.
The marginal log likelihood of a batch of examples is computed in the forward pass, and stochastic gradient descent is used to update the parameters.
We find that the way the likelihood computation is implemented in the forward pass can lead to an orders-of-magnitude difference in training time.
For every example, we need to compute its conditional likelihood for every class based on votes from every PLF.
Naively, this will require three layers of for loops through examples, PLFs, and classes.
We can speed up the computation by expressing the conditional log likelihood computation as a sequence of matrix operations.
This optimization trades space for speed by precomputing intermediate values and taking advantage of vectorized matrix operations.

Let $m$ be the number of instances in one batch, $n$ be the number of PLFs and $k$ be the number of classes.
For each batch we precompute accuracy indicator matrices $\mathbf{AI} \in \{-1,1\}^{m\times n \times k}$ and count matrices $\mathbf{N} \in \mathbb{Z}^{m\times n \times k}$  where entry $\mathbf{AI}_{a,i,j} = 1, \mathbf{N}_{a,i,j} = - \log |N_{i,j}|$ if class $y_j$ is in the label subset output by the $i$-th PLF on the $a$-th example, and $\mathbf{AI}_{a,i,j}=-1,  \mathbf{N}_{a,i,j} = - \log |N_{i,j}^\mathrm{C}|$ otherwise.
We also precompute propensity indicator matrices $\mathbf{PI} \in \{0,1\}^{m \times n}$ where entry $\mathbf{PI}_{a,i}=1$ if the $a$-th instance received a non-abstaining vote (vote is not $\mathcal{Y}$) from the $i$-th PLF.
Let $\mathbf{A} \in \mathbb{R}^{n \times k}$ be the log of the accuracy parameters and $\mathbf{B} \in \mathbb{R}^{n}$ be the log of the propensity parameters.
We can map these parameters back to probability space as
\begin{equation}
\label{eq:acc_and_prop}
\begin{aligned}
\alpha_{i,j} &= \frac{\exp (\mathbf{A}_{i,j})}{\exp(\mathbf{A}_{i,j})+\exp(-\mathbf{A}_{i,j})}
\text{\;\;and\;\;} \\
\beta_b &=\frac{\exp(\mathbf{B}_{i})}{\exp(\mathbf{B}_{i}) + 1} \; .
\end{aligned}
\end{equation}
We extend $\mathbf{PI}$, $\mathbf{A}$, and $\mathbf{B}$ to $\mathbf{PI}^{\text{ext}}$, $\mathbf{A}^{\text{ext}}$, and $\mathbf{B}^{\text{ext}}$ in 3 dimensions, with $\mathbf{PI}$ replicated along the third axis $k$ times, $\mathbf{A}$ replicated along the first axis $m$ times, and $\mathbf{B}$ replicated along the first axis $m$ times and third axis $k$ times.
Then, during each forward pass, we only need to calculate normalizing matrices $\textbf{ZA} \in \mathbb{R}^{n\times k}$ and $\textbf{ZB} \in \mathbb{R}^n$ for accuracy and propensity respectively where
\begin{equation}
\begin{aligned}
        \textbf{ZA}_{i,j} &= - \log \left(\exp(\textbf{A}_{i,j}) + \exp (-\textbf{A}_{i,j}) \right)
\text{\;  and  \;} \\
\textbf{ZB}_i &= - \log(\exp (\textbf{B}_i) + 1) \; .
\end{aligned}
\end{equation}

We similarly extend $\textbf{ZA}$ and $\textbf{ZB}$ to $\textbf{ZA}^{\text{ext}}$ and $\textbf{ZB}^{\text{ext}}$.
During the forward pass we calculate the batch conditional log likelihood by first computing a 3-dimension tensor:
\begin{equation*}
\underset{m\times n\times k}{\mathbf{T}} = \underset{m\times n\times k}{\mathbf{A}^{\text{ext}}}
    \odot \underset{m\times n\times k}{\mathbf{AI}}
    + \underset{m\times n\times k}{\mathbf{N}}
    + \underset{m\times n\times k}{\mathbf{B}^{\text{ext}}}
    + \underset{m\times n\times k}{\mathbf{ZA}^{\text{ext}}}
\end{equation*}
and then summing over the $n$ PLFs:
\begin{equation}
\underset{m\times k}{\log P(G | Y)} = \displaystyle \sum_{n}
\left(
\underset{m\times n\times k}{\mathbf{ZB}^{\text{ext}}}
+ \underset{m\times n\times k} {\mathbf{PI}^{\text{ext}}} \odot 
    \underset{m\times n\times k}{\mathbf{T}} \right)
\end{equation}
where $\odot$ is element-wise multiplication.
The marginal likelihood is then combuted by summing over the $k$ possible classes.
This modification allows us to remove for loops in our code from the computation graph, and instead use only optimized matrix operations.
This approach leads to a $300\times$ speedup in training time compared to a naive approach.
This speedup makes the framework practical for iterative PLF development.
For example, learning with 100k examples and 10 PLFs on an Intel i5-6600k CPU takes one minute.

\paragraph{Identifiability}
\label{sec:ident}
An important theoretical question is whether it is reasonable to try to learn the parameters of $P(G, Y)$ even though $Y$ is never observed.
We answer this question affirmatively by showing that as long as the codomains of the PLFs are sufficiently targeted or diverse, it is possible to determine the parameters of the label model (up to label swapping) using only the distribution of PLF outputs $P(G)$, except for on a measure zero subset of the space of possible parameter values.
This property is the strongest useful notion of identifiability for models with latent variables~\citep{allman:causalinf15}.
A model whose parameters can be determined except for on a measure zero subset is called \textit{generically identifiable}.
\textit{Label swapping} refers to the fact that unobserved classes in a latent variable model can be relabeled without changing the observed distribution.
This means that the map going from the observed distribution of a label model with $k$ classes to parameter values is at best $k!$-to-one and cannot be one-to-one even under ideal conditions.
In practice, label swapping is not an issue because most PLFs are more accurate than random guessing.
We state a condition on the PLF codomains that is sufficient for identifiability in the following theorem.

\begin{theorem}
The parameters of the model P(G, Y) described in Section \ref{section:label_model} are generically identifiable up to label swapping provided that the collection $G$ of partial labeling functions can be partitioned into three disjoint non-empty sets $S_1$, $S_2$, and $S_3$ such that, for sets $j=1,2$ and all classes $y \in \mathcal{Y}$, we can choose label sets $t_i \in T(G_i)$ satisfying 
$\bigcap_{G_i \in S_j} t_i = \{y\}$.
\end{theorem}

The proof is given in Appendix \ref{section:proof_theorem1}.
This theorem tells us that it is reasonable to try to estimate the PLF accuracies even though the true class labels are never observed.
Our proof adapts ideas presented in Theorem 4 of \citet{allman:annalstat09}, which uses Kruskal's unique factorization theorem and feature grouping to establish conditions for the generic identifiability of a naive Bayes model with arbitrary parameters.
Since the space of models we consider is equivalent to a measure zero subset of the parameters in an arbitrary naive Bayes model, an additional proof is needed to show that these parameters are generically identifiable.
We develop a novel argument to show that the above is a sufficient condition for identifiability.
We show that for any distribution satisfying the condition described in Theorem 1, we can construct matrices representing factors of the joint distribution over observations that generically have full Kruskal rank.
This is sufficient to satisfy the conditions in Kruskal's theorem.

In words, the condition described in Theorem 1 requires that for each class $y$, we can select a label group from the codomain of each PLF in $S_1$ such that the intersection of these label groups contains only the class $y$.
This condition also applies to $S_2$.
One way to satisfy this condition is to create PLFs that produce single-class label groups.
For example, if PLF $G_i$ contains $\{1\}, \{2\},...,\{k\}$ in its codomain, then any set $S_j$ that contains $G_i$ will satisfy the Theorem 1 condition. However, even if no PLFs output any single-class label sets, it is still possible for the label model parameters to be identifiable because the condition can also be satisfied by using multiple PLFs with different codomains. Suppose that we want to show that the condition is satisfied for class 1 and we have $\{1,2,3\} \in T(G_1)$, $\{1,3,4\} \in T(G_2)$ and $\{1,2,4\} \in T(G_3)$.
The intersection of these sets is $\{1\}$.

\subsection{Noise-Aware Classifier}
\label{section:end_model}
The final stage of our framework is to train a classifier.
After $P(G,Y)$ is estimated with unlabeled data, we compute the posterior $P(Y|G)$.
Then, we minimize the expected empirical risk with respect to this distribution.
For classifiers that output probabilistic predictions, the loss function becomes the cross-entropy loss weighted by the posterior over true labels.
As in other PWS frameworks~\citep{ratner:neurips16,ratner:vldbj20,ratner:aaai19,safranchik:aaai20}, many off-the-shelf neural networks can be chosen based on the task.

%% file: sections/experiments.tex
\section{EXPERIMENTAL RESULTS}
\label{sec:exp}
We demonstrate benefits of incorporating partial labels into PWS on applications in text and object classification.
In Section~\ref{sec:text}, we compare our framework with baselines that (1) use only traditional labeling functions and (2) heuristically aggregate partial labels without a probabilistic model.
Our proposed approach significantly improves accuracy over both baselines.
In Section~\ref{sec:image}, we use pretrained visual attribute detectors as PLFs for classifying unseen objects.
Our framework achieves accuracy that is competitive with recent embedding-based transductive ZSL methods.
While our framework is not designed specifically for ZSL, we present this comparison to demonstrate its flexibility and show another scenario where modeling the noise of multiple partial labeling functions can significantly improve performance relative to a heuristic approach.
It shows that  discrete attribute detectors can be competitive with recent ZSL approaches.
These two sets of experiments are complementary, showing that our framework benefits both hand-engineered rules and partial labeling schemes defined in prior work.

We make available the code for our framework\footnote{\href{https://github.com/BatsResearch/nplm}{\texttt{github.com/BatsResearch/nplm}}} and experiments.\footnote{\href{https://github.com/BatsResearch/yu-aistats22-code}{\texttt{github.com/BatsResearch/yu-aistats22-code}}}
Additional details  about the experiments, datasets, and methods are available in Appendix~\ref{sec:ed}.

\input{figs/tables/text_tasks}
\subsection{Text Classification}
\label{sec:text}

We first evaluate the benefit of incorporating PLFs into text classification with hand-engineered rules.

\paragraph{Datasets}
We consider three datasets.
First, SciCite \citep{cohan:naacl19} is a citation classification dataset sourced from scientific literature.
The corresponding task is to classify a citation as referring to either background information, method details, or results.
Second, TREC-6 \citep{li:coling02} is a question classification dataset containing open-domain questions.
The task is to classifiy each question as asking about one of six semantic categories.
Finally, AG-News \citep{gulli:web05} is a large-scale news dataset.
The task is to classify each example as one of four topics.

\paragraph{PLF Development}
\input{figs/plf_pseudocode}

For text tasks, we develop PLFs by inspecting examples from the development set for each dataset (916 examples for SciCite, 389 for TREC-6, and 500 for AG-News).
We implement heuristic rules as Python functions that take as input the example text and any available metadata (such as the name of the section in which the sentence appears for SciCite).
Most rules rely on checking for keywords or other surface patterns in the text.
Figure \ref{fig:plf_example} shows an example for TREC-6, in which we use the first word of a question to vote on what type of question it is.
This example illustrates some of the utility of PLFs.
Some words like ``who'' are sufficient to reliabily identify a single class.
Others like ``how'' greatly narrow the set of possible lables, even though they do not specify a single label.
The full set of PLFs are in Appendix \ref{sec:plf} and the experiment code.

\paragraph{Methods}
We evaluate methods for aggregating the outputs of the PLFs to train an end model.
First, as a baseline, we consider using only the PLFs that are equivalent to traditional labeling functions, i.e., they always output one label or abstain. We call this baseline \textbf{LFs Only}.
Second, as another baseline we use a heuristic called \textbf{Nearest Class (NC)}, which chooses the class with the highest number of compatible partial labels.
This baseline is a generalized majority vote heuristic for PLFs.
Finally, our method, called \textbf{Noisy Partial Label Model (NPLM)}, is our label model from Section~\ref{section:label_model}.
In all cases, we use the estimated labels to fine-tune a pretrained BERT~\citep{devlin:naacl19} base uncased English model with a three-layer classification head.
Following prior work~\citep{ratner:neurips16, ratner:vldbj20}, we use the expected cross entropy w.r.t.~$P(Y|G)$ as our training loss.
Hyperparameters and additional end model details are in Appendix \ref{sec:ed}.
As ablations, we also report performance using the aggregated PLF outputs directly as predictions, without training an end model, denoted \textbf{(w/o End)}.

\paragraph{Results}
We report mean micro-averaged accuracy and macro-averaged F1 of the compared methods in Table \ref{text_res} on the standard test sets.
Results using the end model are shown with 95\% confidence intervals obtained using five different random seeds.
NPLM consistently improves F1 and accuracy relative to LFs Only (8.6 and 8.1 percentage points on average, respectively) and NC (8.5 and 12.8 percentage points on average, respectively).
The performance advantage over LFs Only demonstrates the benefits of additional weak supervision that can be expressed as PLFs, and the advantage over NC demonstrates that the proposed label model is learning useful information.
The ablated versions of the methods significantly underperform their counterparts, showing that in all cases the end model learns to generalize beyond the information contained in the weak supervision heuristics.
Many of the errors are on examples for which all supervision sources abstain, where a label is chosen arbitrarily or according to the class prior $P(Y)$.
For context, we also report the performance of the end model trained on the development set with ground-truth labels.
NPLM significantly outperforms it in most cases.
On TREC-6, the supervised baseline has a higher accuracy but much lower macro-averaged F1, indicating that our method does significantly better on the rarer classes.

\input{figs/tables/zsl_combined}
\subsection{Object Classification}
\label{sec:image}

In this task, we show how our framework can be used to model discrete visual attribute detectors, and that this approach can achieve results competitive with recent embedding-based ZSL methods.
Although they have not been used often in recent ZSL work, discrete attribute detectors have benefits such as modularity and interpretability.
These experiments show that modeling them as PLFs with our unsupervised label model can lead to good accuracy.

\paragraph{Datasets}
We consider the Large-Scale Attribute Dataset (LAD)~\citep{zhao:cvpr-ws19} and Animals with Attributes 2 (AwA2)~\citep{xian:pami18}, which both provide class-level discrete visual attributes.
LAD is a recently proposed attribute-based dataset with 78k instances that organizes common objects into five sub-datasets: electronics, vehicles, fruits, hairstyles and animals.
For each sub-dataset, the classes are divided into five folds of seen and unseen classes, and average performance over all tasks is used as a benchmark for ZSL.
AwA2 is a popular ZSL animal classification dataset consisting of $\sim$30k instances with 85 binary attributes, 40 seen classes, and 10 unseen classes.

\paragraph{PLF Development}
Following early work on zero-shot object classifcation~\citep{farhadi:cvpr09, lampert:cvpr09, palatucci:neurips09,jayaraman:neurips14}, we model each visual attribute in the datasets with a binary classifier.
In all cases, the classifiers are trained on the seen classes for that task or fold, and the unseen classes are not used at all, not even as validation data.
To create classifiers for LAD, we extract features from a ResNet-50~\citep{he:cvpr16} pretrained on ILSVRC~\citep{russakovsky:ijcv15}, in order to compare fairly with prior work.
For AwA2, we fine-tune a pretrained ResNet-101 on the seen classes.
Each class is trained with respect to the class-wise attribute annotations on the training sets of the seen classes.
We define the PLFs according to the provided attribute annotations from the respective datasets.

\paragraph{Methods}
We incorporate PLFs into the NC and NPLM methods as described in Section~\ref{sec:text}.
We use examples of the unseen classes as our unlabeled data.
In all cases, our end models are three-layer perceptrons trained on the extracted features (ResNet-50 for LAD and ResNet-101 for AwA2).
As with the text data, we use the expected cross entropy with respect to $P(Y|G)$ as our training loss.
Following the literature, for LAD we evaluate in a strict zero-shot setting, meaning that the model is only evaluated on unseen classes.
For AwA2, we evaluate in a generalized zero-shot setting, meaning that the model is evaluated on both seen and unseen classes, so we mix the unseen classes with estimated labels and seen classes with given labels during training.
Again, additional details are in Appendix \ref{sec:ed}.
We compare with three recent transductive, embedding-based ZSL methods: QFSL~\citep{song:cvpr18}, VCL~\citep{wan:neurips19}, and WDVSc~\citep{wan:neurips19}.
For context, we also report results from three standard inductive methods, ConSE~\citep{norouzi:conse2013}, ESZSL~\citep{romera:icml15}, SynC~\citep{changpinyo:cvpr16}, although they are at a disadvantage because they do not access the unlabeled data nor any information about the unseen classes.
For LAD, we replicate and report the results of WDVSc and VCL using the same features.

\paragraph{Results}
We report the average results and 95\% confidence intervals based on five random seeds in Table~\ref{image_res}.
Similar to the text classification tasks, NPLM significantly outperforms NC (an average of 9.8 percentage points on LAD and 21.5 percentage points on AwA2), and the ablations show that the end model generalizes beyond the PLFs.
NPLM is also competitive with WDVSc, the top-performing ZSL method, either slightly underperforming or outperforming.

%% file: figs/tables/text_tasks.tex
\begin{table*}[ht!]
  \centering
   \resizebox{0.75\textwidth}{!} {
  \begin{tabular}{lcccccc}
    \toprule
    &
    \multicolumn{2}{c}{SciCite}        &       
   \multicolumn{2}{c}{TREC-6}          & 
    \multicolumn{2}{c}{AG-News}\\
    \cmidrule(r){2-3} \cmidrule(r){4-5}  \cmidrule(r){6-7} 
    &ACC& F1 &ACC & F1 &ACC& F1\\
    
    \midrule

     Supervised (Dev. Set)  & 78.5$\pm$1.1 & 75.5$\pm$1.5 & 88.3$\pm$1.6 & 76.2$\pm$2.0 & 80.1$\pm$1.7& 79.9$\pm$1.9 \\
 \midrule
     LFs Only (w/o End)  &  65.1 & 44.7 & 22.0 & 23.8 & 33.0 & 25.1  \\
     NC (w/o End)       & 73.2 & 69.2 & 29.6 & 32.6 & 46.1 & 43.5  \\
     NPLM (w/o End)      & 71.5 & 69.4  & 38.2 & 43.0 & 51.1 & 49.4 \\

 \midrule

    LFs Only  & 78.6$\pm$0.7 & 76.7$\pm$0.8 & 67.2$\pm$0.9 & 69.3$\pm$1.0 &79.8$\pm$0.9&  79.6$\pm$0.8 \\

    NC   & 80.2$\pm$0.6 & 77.7$\pm$0.6 & 67.8$\pm$1.5 & 56.5$\pm$0.3 & 78.0$\pm$1.0 & 77.2$\pm$1.0  \\

    NPLM   & 81.4$\pm$1.3 & 79.5$\pm$1.3 & 85.0$\pm$0.8 & 85.7$\pm$0.7 & 85.0$\pm$0.5& 84.7$\pm$0.5  \\
   \midrule

    NPLM vs. LFs Only &	$\uparrow$ 2.8 & $\uparrow$ 2.8 &	$\uparrow$ 17.8 & 	$\uparrow$ 16.4 &	$\uparrow$ 5.2&	$\uparrow$ 5.1\\
    NPLM vs. NC &	$\uparrow$ 1.2 & $\uparrow$ 1.8 &	$\uparrow$ 17.2 & 	$\uparrow$ 29.2&	$\uparrow$ 7.0 &  $\uparrow$ 7.5\\
    \bottomrule
  \end{tabular}
  }
    \caption{Results for text classification with mean accuracy (ACC), macro F1 (F1) and 95\% CIs.}
 \label{text_res}
\end{table*}

%% file: figs/plf_pseudocode.tex
\begin{figure}[t]
\begin{minted}
[
framesep=1mm,
baselinestretch=0.4,
fontsize=\footnotesize
]{python}
def first_word_PLF(instance):
    '''
    Label by first word of question.
    
    ABBR - Abbreviation
    DESC - Description and concepts
    ENTY - Entities
    HUM - Human beings
    LOC - Locations
    NUM - Numeric values
    '''
    word = instance.split()[0].lower()
    if word == "who": return [HUM]
    elif word == "where": return [LOC]
    elif word == "when": return [NUM]
    elif word == "why": return [DESC]
    elif word == "how": return [DESC, NUM]
    elif word == "name": return [ENTY, HUM]
    else: return [ABBR, DESC, ENTY, 
                  HUM, LOC, NUM]  # Abstain
        
\end{minted}
\caption{A partial labeling function developed for the TREC-6 question-type task.
It uses the first word of the sentence to possibly narrow down the set of labels.}
\label{fig:plf_example}

\end{figure}
\vspace{-5pt}

%% file: figs/tables/zsl_combined.tex
\newcommand{\mcau}{$U$}
\newcommand{\mcas}{$S$}

\begin{table*}[ht!]  
  \centering
   \resizebox{0.95\textwidth}{!} {
  \begin{tabular}{lccccccccc}
    \toprule &
     \multicolumn{6}{c}{LAD (ACC)} &       
   \multicolumn{3}{c}{AwA2 (MCA)} \\
   \cmidrule(r){2-7} \cmidrule(r){8-10}  
     & Animals & Fruit & Vehicles &  Electronics& Hairstyles & Avg. & \mcau & \mcas &  $\mathcal{H}$ \\
    \midrule
    ConSE---\cite{norouzi:conse2013} & 36.9 & 29.8 & 37.5 & 28.3 & 24.6 & 31.4 &0.5&90.6&1.0\\
    ESZSL---\cite{romera:icml15} & 50.2 & 37.2& 45.8& 32.8& 31.8& 39.6&77.8 &5.9& 11.0\\
    SynC---\cite{changpinyo:cvpr16} & 61.6 & 51.4& 54.9 & 43.0 & 29.1& 48.0 &90.5 &10.0& 18.0\\
    \midrule
    
    VCL---\cite{wan:neurips19} & 75.4$\pm$ 0.8&35.0$\pm$ 1.0&62.4$\pm$ 0.5&36.7$\pm$ 0.5&33.8$\pm$ 0.7&48.7$\pm$ 0.3 &21.4&89.6&34.6\\
    
    QFSL---\cite{song:cvpr18} &-&-&-&-&-&-&66.2&93.1&77.4\\
    
    WDVSc---\cite{wan:neurips19} & 
97.2$\pm$ 0.8&43.3$\pm$ 1.3&82.1$\pm$ 0.6&54.8$\pm$ 1.1&31.1$\pm$ 2.6&61.7$\pm$ 0.6 &76.4&88.1&81.8\\
    \midrule
    NC (w/o End) & 65.8 & 31.2 & 60.3& 40.3& 39.1& 47.3 & 47.7 &- &-\\
    NPLM (w/o End) & 86.0 &
38.7 &
73.5 &
51.8 &
45.9 &
59.2 & 68.2 & - & -\\

    \midrule
     NC & 71.9$\pm$1.2&36.2$\pm$0.6&65.3$\pm$1.2&48.0$\pm$0.7&40.9$\pm$0.5&52.5$\pm$0.3& 43.1$\pm$ 1.2&91.8$\pm$ 0.2&58.6$\pm$ 1.1\\
    NPLM   &
87.6$\pm$ 0.2&42.4$\pm$ 0.8&77.0$\pm$ 0.2&57.7$\pm$ 0.7&46.9$\pm$ 0.9&62.3$\pm$ 0.2 & 
71.1$\pm$ 0.6 &91.9$\pm$ 0.1&80.1$\pm$ 0.3\\

    \midrule
    NPLM vs. NC &	$\uparrow$ 15.7 & $\uparrow$ 6.2&	$\uparrow$ 11.7 & 	$\uparrow$ 9.7&	$\uparrow$ 6.0&	$\uparrow$ 9.8&	$\uparrow$ 28.0&	-&	$\uparrow$ 21.5\\
    \bottomrule
  \end{tabular}
  }

  \caption{Results for object classification.
  For LAD, we report mean accuracy (ACC) with 95\% CIs across the five standard splits for each of the five subtasks.
  For AWA2, we report mean class accuracy (MCA) with 95\% CIs.
  We evaluate AWA2 in a generalized setting: $S$ and $U$ denotes MCA on the seen and unseen classes respectively. $\mathcal{H}$ is the harmonic mean.}
  \label{image_res}
\end{table*}

%% file: sections/conclusion.tex
\section{CONCLUSION}
\label{sec:conc}

We have introduced a new capability for programmatic weak supervision (PWS): the ability to learn from partial labeling functions using a novel probabilistic label model.
We demonstrated a scalable way to learn these models, and our theoretical analysis shows they are generically identifiable up to label swapping, the strongest useful notion of identifiability for latent variable models~\citep{allman:causalinf15}.
Our experiments show that our framework can (1) significantly improve the accuracy of PWS on text classification tasks and (2) enable pre-trained attribute detectors to achieve performance comparable to embedding-based methods for transductive ZSL on object classification tasks.
We aim to enable the incorporation of a wider range of supervision sources into PWS systems.
As future work, we envision creating libraries of rules and pre-trained models that are more generic and modular because they are freed from the requirement that they narrow the label space down to a single class.

%% file: sections/acknowledgements.tex
\section*{Acknowledgements}
This material is based on research sponsored by Defense Advanced Research Projects Agency
(DARPA) and Air Force Research Laboratory (AFRL) under agreement number FA8750-19-2-1006.
The U.S. Government is authorized to reproduce and distribute reprints for Governmental purposes
notwithstanding any copyright notation thereon. The views and conclusions contained herein are
those of the authors and should not be interpreted as necessarily representing the official policies or
endorsements, either expressed or implied, of Defense Advanced Research Projects Agency (DARPA)
and Air Force Research Laboratory (AFRL) or the U.S. Government. We gratefully acknowledge
support from Google and Cisco. Disclosure: Stephen Bach is an advisor to Snorkel AI, a company
that provides software and services for weakly supervised machine learning.

%% file: sections/appendix.tex
\appendix
\section{LIMITATIONS AND BROADER IMPACTS}

Our work expands the space of supervision sources that can be incorporated into PWS systems.
Weak supervision is complementary to many other techniques, such as semi-supervised learning~\citep{chapelle:book09, van:ml20}, transfer learning~\citep{pan:tkde09, zhuang:ieee20}, active learning~\citep{cohn:jair96, settles:lectures12}, and zero-shot data generation~\citep{bucher:cvpr17, verma:cvpr18, felix:eccv18, sariyildiz:cvpr19,xian:cvpr19,narayan:eccv20}.
A limitation of our work is that exploring how partial labeling functions interact with these techniques is left as future work.
The same is true for complementary techniques within weak supervision, such as adversarial label learning~\citep{arachie:jmlr21}, learning rules from labeled exemplars~\citep{awasthi:iclr20}, and weak supervision with self-training~\citep{karamanolakis:naacl21}.
Additionally, while PWS can enable more rapid development, its dependence on heuristics introduces the potential for bias.
For this reason, auditing any created models for potential negative impacts is as important, if not more important, in PWS as in traditional supervised learning~\citep{saleiro:arxiv18, mehrabi:arxiv19}.

\section{PROOF OF THEOREM 1} \label{section:proof_theorem1}
Theorem 1 provides sufficient conditions for the generic identifiability of the label model described in Section \ref{section:label_model}.
In this section, we prove this theorem. 
Our proof is non-trivial because our label model yields probability distributions that are a measure-zero subset of the distributions considered by \citet{allman:annalstat09}.
\citet{allman:annalstat09} allows each entry of the class-conditional distributions to be any value in the interval $[0, 1]$ such that the entries sum to 1, whereas our label model imposes additional algebraic constraints on the entries.
\citet{allman:annalstat09} establishes identifiability except for on a measure-zero subset of the distributions that they consider, but we are unable to directly apply their results because our family of distributions might be contained in the measure-zero subset they exclude. 
It is therefore necessary to establish that the set of distributions for which identifiability does not exist is of measure zero with respect to the distributions that can be produced by our label model, or, equivalently, the set of values of the accuracies $\alpha_{i,j}$, propensities $\beta_{i}$, and class balance $P(Y)$ for which identifiability does not exist has measure zero with respect to the set of all possible parameter values. 

\paragraph{Background} The key tool that we use in our proof is Kruskal's unique factorization theorem, which relies on the concept of Kruskal rank \citep{kruskal:linalg77}. The \textit{Kruskal rank} of a matrix is defined to be the largest integer $n$ such that every set of $n$ rows is linearly independent. A useful fact is that a matrix with full row rank also has full Kruskal rank. Kruskal's theorem says that if, for $u=1,2,3$, we have a $k \times r_u$ matrix $M_u$ with Kruskal rank $I_u$, and these $I_u$ satisfy 
\begin{equation}
    I_1 + I_2 + I_3 \geq 2k+2
\end{equation}
then, given only the three-dimensional tensor $M$ where entry $(a,b,c)$ is given by 
\begin{equation} \label{eq:tensor_M}
    M(a,b,c) = \sum_{j=1}^k M_{1}(j,a)M_{2}(j,b)M_{3}(j,c)
\end{equation}
we can recover the original matrices $M_u$.

 \citet{allman:annalstat09} makes the connection that the probability distribution of a latent variable model with three observed variables and one latent variable that takes on a finite set of values can be described by the matrix $M$. Each $M_u$ can be intrepreted as a conditional probability matrix where row $c$ is a probability distribution over the possible values of feature $u$ given that the latent variable has value $c$. 

In situations where there are more than three observed variables, variables can be combined to form ``grouped" variables that satisfy the theorem conditions, if needed. In our case, it is possible that individual PLFs do not have codomains that are large and diverse enough to satisfy the Kruskal rank requirement. In these situations, the condition can be satisfied by amalgamating multiple PLFs to form a grouped PLF, which can be viewed as an observed variable with a codomain that is the Cartesian product of the codomains of its member PLFs. 

We show that the conditions of Theorem 1 ensure that, for a generic choice of parameters in a $k$-class model, there is a tripartion of the PLFs such that two of the corresponding conditional probability matrices have full Kruskal rank ($k$) and the third conditional probability matrix has a Kruskal rank of at least 2. Thus, the conditions of Kruskal's unique factorization theorem are satisfied, so we can recover the conditional probability matrices, from which the accuracies $\alpha_{i,j}$, propensities $\beta_{i}$, and class balance $P(Y)$ can be computed by solving a system of equations.

\begin{proof}[Proof of Theorem 1] $S_1$, $S_2$, and $S_3$ partition the PLFs into three disjoint subsets. For $u=1,2,3$, define some ordering to the PLFs in subset $S_u$ so that $S_{u,i}$ gives the $i$-th PLF in the subset. We will treat $S_u$ as a ``grouped" PLF with codomain $\mathcal{G}(S_u) = \{(t_1, t_2, ..., t_{|S_j|}) \mid t_1 \in \mathcal{G}(S_{u,1}), t_2 \in \mathcal{G}(S_{u,2}), ..., t_{|S_u|} \in \mathcal{G}(S_{u,|S_u|})\}$, where $\mathcal{G}(G)$ denotes the codomain of PLF $G$. Let $M_u$ denote the $k \times |\mathcal{G}(S_u)|$ conditional probability matrix for the combined output of all PLFs in subset $S_u$, where each entry is a product containing some combination of $\beta_{i}$, $(1-\beta_{i})$, $\alpha_{i,j}$, $(1-\alpha_{i,j})$, and normalizing constants. 
We assume that the class balance $P(Y)$ has positive entries and all PLFs have non-zero propensities $\beta_{i}$, because any extraneous class labels or non-voting PLFs would be removed. Define $\tilde{M}_1=\mathsf{diag}(P(Y)) M_1$, where $\mathsf{diag}(\mathbf{v})$ denotes the matrix with the entries of vector $\mathbf{v}$ along its main diagonal and zeros elsewhere. 
$P(G$), the observed distribution of PLF outputs, corresponds to the three-dimensional tensor obtained from applying Equation \ref{eq:tensor_M} to $\tilde{M}_1$, $M_2$, and $M_3$.
We will consider the Kruskal ranks of $\tilde{M}_1$, $M_2$, and $M_3$, which we respectively denote $I_1$, $I_2$, and $I_3$.

We first consider $M_2$. The (row) rank of a matrix $A$ is equal to the largest integer $n$ for which there exists an $n \times n$ submatrix of $A$ that has a nonzero determinant. 
The determinant of such a submatrix is called an $n$-minor. $M_2$ has less than full row rank if and only if all of its $k$-minors are zero. 
This condition can be expressed as the nonvanishing of a polynomial in the entries of $M_2$, which are themselves functions of the label model parameters.
In other words, the set of parameter values for which $M_2$ does not full row rank is the zero set of this polynomial. As described in \citet{allman:annalstat09}, so long as the polynomial is not identically zero, the parameter values yielding less than full row rank is a measure-zero subset of the full parameter space. 
To show that this polynomial is not zero for all values in the parameter space, it is sufficient to show that there exists at least one set of parameter values for which the polynomial is nonzero, or, equivalently, that there is a set of parameter values for which $M_2$ has full row rank. 

The values of the propensities $\beta_{i}$ and class balance $P(Y)$ do not affect row rank as long as they are nonnegative, as assumed above.
We now show that there is a setting of the accuracies $\alpha_{i,j}$ for which the Kruskal rank of $M_2$ is $k$. 
Set all $\alpha_{i,j}=1$. 
By the conditions of Theorem 1, for each class $c$, there is an output in the codomain of $S_2$ for which $c$ appears in all of the individual PLF outputs and no other class appears in all outputs. 
This implies the following two statements about the column in $M_2$ that is associated with this output: (1) the $c$-th entry of this column does not contain $(1-\alpha_{i,j})$ in its product, and (2) all other entries are products containing at least one $(1-\alpha_{i,j})$.
When $\alpha_{i,j}=1$, these entries containing $(1-\alpha_{i,j})$ are all zero.
In other words, $M_2$ has $k$ columns that are all zero except for a single entry, and the row containing this entry is different across the $k$ columns. 
These columns form the basis of a column space of dimension $k$. 
For any matrix $A$, dim(Col $A$) = dim(Row $A$) = row rank of $A$. Thus, the row rank of $M_2$ is $k$. Since $M_2$ has full row rank when all $\alpha_{i,j}=1$, it also has full Kruskal rank. This shows that the polynomial whose nonvanishing determines whether $M_2$ has full row rank is not identically zero, so $M_2$ generically has full row and Kruskal rank. 

We now consider $\tilde{M}_1$. The arguments applied to $M_2$ can be applied exactly to $M_1$, but we are interested in $\tilde{M}_1=\mathsf{diag}(P(Y)) M_1$. 
However, since we assumed that $P(Y)$ contains only positive entries, and multiplying each row of a matrix by a nonzero scalar does not change its row rank, the same arguments can be applied to $\tilde{M}_1$. 
We conclude that $\tilde{M}_1$ also generically has full Kruskal rank. 

Finally, we consider $M_3$. The Kruskal rank of a matrix is less than two only if there are two rows that are scalar multiples of each other. 
This can happen in our model only when the class conditional accuracies for two classes are exactly equal, which corresponds to a measure zero subset of the parameter space. 
Thus, we can generically assume that $M_3$ has a Kruskal rank of $I_3 \geq 2$.

Since, generically, $M_1$ and $M_2$ have Kruskal ranks of $k$ and $M_3$ has a Kruskal rank of at least 2, we have $I_1+I_2+I_3 \geq 2k + 2$, so Kruskal's unique factorization theorem tells us that we can recover $\tilde{M_1}$, $M_2$, and $M_3$ from $P(G)$, the observed distribution of PLF outputs. Once $\tilde{M_1}$, $M_2$, and $M_3$ are known, the accuracies $\alpha_{i,j}$, propensities $\beta_{i}$, and class balance $P(Y)$ can be computed using algebraic manipulations. 
\end{proof}

\section{DATASET INFORMATION}
\label{sec:data}
\textbf{SciCite} \citep{cohan:naacl19} is a citation purpose classification dataset containing 8243 train, 916 development and 1861 test instances of 3 categories sourced from scientific literatures and it is publicly available under Apache License 2.0.

\textbf{TREC-6} \citep{li:coling02} is a publicly available dataset for research use and it is a question classification dataset containing a broad amount of open-domain questions from 6 semantic categories. Since the original dataset lack a validation/development set, we sample 389 instances from the training set to make a train/dev/test size split of 5063/389/500 respectively.

\textbf{AG-News} \citep{gulli:web05, zhang:neurips15} is a publicly available dataset for research use. It is a large-scale news topic classification dataset containing 4 categories. We similarly sample 500 training instances as our development set.
The train/dev/test size are 119.5k/500/7600 respectively.

\textbf{LAD} \citep{zhao:cvpr-ws19}  is a publicly available dataset for research use. It has approximately 78k instances that organizes common objects into five sub-datasets: electronics, vehicles, fruits, hairstyles and animals. Each sub-dataset is associated with 5 different seen/unseen class splits.

\textbf{AwA2} \citep{xian:pami18} is a publicly available dataset for research use. It has 85 binary attributes for 50 animal classes. For our experiment, following the dataset authors, we adopt the proposed split that divides the 50 classes into 40 seen classes and 10 unseen classes.

All datasets we use are publicly available standard research datasets.
These datasets generally do not contain personally identifiable information.
Public figures are sometimes mentioned in the text datasets.

\section{Additional Experiment Details}
\label{sec:ed}

For the PLFs development and label modeling stage of the text classification task, the experiments are set on a local PC with Intel i5-6600k CPU and 32 GB of RAM. For the discriminative modeling and PLFs development that involves neural network inference/training for the object classification task, we perform our experiments on virtual computing instances with Intel Xeon E5-2698 v4 CPU, 1 NVIDIA V100 GPU or 1 NVIDIA RTX 3090 GPU and 32 GB of RAM.

\subsection{Text Classification}
\label{sec:edt}

For both LFs Only and NPLM, following prior practices in programmatic weak supervision \citep{ratner:vldbj20, bach:sigmod19-industrial}, we filter the training instances by only retaining ones with at least one PLFs/LFs votes and the filtered instances are used for LF-Only/NPLM label and end model training.
For the optimization of LFs Only and NPLM, we use a initial learning rate of 0.01 and a reduce-learning-rate-on-plateau learning rate scheduler with decreasing factor of 0.1.
We train the NPLM/LFs Only Label models for 5 epochs.

For the end model, we adopt a pretrained(English) BERT-base model \citep{devlin:naacl19} (bert-base-uncased) with a 3-layer classification head. The model is implemented with AllenNLP \citep{gardner:AllenNLP17}. The 3-layer classification head has a hidden size of dimension 256 with LeakyReLU as activation, batch normalization and a dropout layer with 50\% probability. For all of the end models, we use AdamW (ADAM with weight decay) \citep{loshchilov:iclr19} optimizer and we train the models with a 16 training batch size. For AG-News, we train 20 epochs and a starting learning rate of 5e-7 and we set the gradient clipping threshold to 2.0. For TREC-6, we train with a total of 25 epochs with a 2.0 gradient clip and 3e-5 initial learning rate. For SciCite, we train the model with a total of 20 epochs with a 2.0 gradient clip and 3e-6 initial learning rate. The best end model is picked based on the best validation macro-averaged F1. Please refer to $<$dataset$>$\_pipeline.ipynb, run\_end\_model\_$<$dataset$>$.py, and /end/backbones/text\_classifiers.py in the experiment code repository for corresponding code.\footnote{\href{https://github.com/BatsResearch/yu-aistats22-code}{\texttt{github.com/BatsResearch/yu-aistats22-code}}}

\subsection{Object Classification}
\label{sec:edzsl}

We use AwA2 for our generalized zero-shot experiments and the sub-tasks of LAD for zero-shot evaluations.
For AwA2, we follow the \textit{proposed splits} and we  follow the seen/unseen class split guide noted by the original authors for LAD.
We adopt previous practices and guidelines in evaluating generalized AwA results, using average per class top-1 accuracy (or mean class accuracy, MCA) as the main performance metric for both unseen and seen classes at test and then report the harmonic mean. 
For LAD, we follow the authors’ practices to report the average accuracy over the 5 sub-categories, each with 5 different seen/unseen class splits. 

While the train/validation split among the seen classes are given in AwA2, LAD does not supply a train/validation split among the seen classes.
We randomly sample at least one and at most 10\% of the seen classes as validation classes for the detector and the rest seen data as training.

For both AwA2 and LAD, we train detectors for each attribute with a 3 layer MLP.
We consider setting the hidden dimensions to either 512 or 1024.
We select the size that gives the higher minimum per-class accuracy.
(In other words, whichever improves the worst-scoring class the most.)
We use ILSVRC-pretrained ResNet-50 features for LAD and seen class-finetuned ResNet-101 features on AwA2. The minority class is balanced by oversampling. We also apply batch normalization and 50\% dropout during training at each layer. The activation function used is LeakyReLU. For the optimization, we adopt a Adam optimizer with initial learning rate from 1e-4 and a multi-step learning rate scheduler. We train the detector with respect to \{100, 300, 500\} epochs with a learning rate scheduling step size of \{30, 80, 200\} respectively. Best model is selected based on best validation accuracy measured on the held-out seen classes.

For the NPLM label model, we use Adam optimizer with a reduce-learning-rate-on-plateau learning rate scheduler.
The full set of hyperparameters can be found in the experiment code repository.
The end model is a 3-layer MLP with both hidden layers size being 1024. We apply batch normalization and 50\% dropout at each layer and it is activated with LeakyReLU. We optimize the end discriminative model with a initial learning rate of 1e-4 and we adopt a reduce-learning-rate-on-plateau learning rate scheduler with decreasing factor of 0.1. For the generalized task on AwA2, we train the model for 11 epochs. For LAD, we pick the best model with the lowest training loss. Same as the text tasks, the training objective is soft cross entropy.

\section{PLFs for Text Classification}
\label{sec:plf}

In this section, we list the detailed partial labeling functions involved in our experiments.
Corresponding code for PLFs can be also found in the supplementary code as ipython notebook files.

For text experiments, for faster and more convenient PLFs development, we developed an abstraction class \textbf{WeakRule}(exec\_module, label\_maps) for each PLFs, which has two crucial arguments/elements: exec\_module that is the decision function programmatically defined and label\_maps that map the result from the decision function to a partial label group.

To make development process of some rules even simpler, we develop a further abstraction inherited from class \textbf{WeakRule}, \textbf{BinaryRERule}. \textbf{BinaryRERule} will exam if specified regular expression match a given sentence and give either positive/negative feedback for each partial label groups or positive/abstain feedback if it is a unipolar Binary PLF (meaning that it only cast supervision signals onto a single partial label group or abstains).\\
\noindent
\textbf{For SciCite} (10 PLFs):\\
Label Index Mapping:\\ 0 - Background 1 - Method 2 - Result\\
\input{figs/plf_sep_sci}
\noindent
\textbf{For TREC-6} (16 PLFs):\\
Label Index Mapping:\\ 0 - Abbreviation 1 - Description and concepts 2 - Entities 3 - Human beings 4 - Locations 5 - Numeric values

\input{figs/plf_sep_trec}

\noindent
\textbf{For AG-News} (11 PLFs):\\
Label Index Mapping: 0 - World 1 - Sports 2 - Business 3 - Science/Technology
\input{figs/plfs_sep_ag}

%% file: figs/plf_sep_sci.tex
\begin{minted}[framesep=1mm,
baselinestretch=0.4,
fontsize=\small,
]{python}
def df1(instance):
    sen = instance['preprocessed_string'].lower()
    if ' we ' in sen 
        or ' our ' in sen 
        or ' by us ' in sen:
        return 1
    return -1

firstperson_rule = WeakRule(
    exec_module=df1, 
    label_maps={0:[0], 1:[1,2]})
\end{minted}
\begin{minted}[framesep=1mm,
baselinestretch=0.4,
fontsize=\small
]{python}
def sectionTitleRule(instance):
    results_pat = 'result|discussion|Lconclusion|observation'
    method_pat = 'method|approach|experiment|evaluation'
    intro_pat = 'introduction|background'
    title = str(instance['sectionName']).lower()
    if re.search(method_pat, title):
        return 1
    elif re.search(intro_pat, title):
        return 0
    elif re.search(results_pat, title):
        return 2
    return -1
sectionTitle_rule = WeakRule(
    exec_module=sectionTitleRule, 
    label_maps={0:[0], 1:[1], 2:[0, 2]})
\end{minted}
\begin{minted}[framesep=1mm,
baselinestretch=0.4,
fontsize=\small
]{python}

def df2(instance):
    sen = instance['preprocessed_string'].lower()
    if ' result ' in sen 
       or ' results ' in sen:
        return 1
    return 0
result_rule = WeakRule(
    exec_module=df2, 
    label_maps={0:[0,1], 1:[2]})

\end{minted}
\begin{minted}[framesep=1mm,
baselinestretch=0.4,
fontsize=\small
]{python}
def length_citation(instance):
    m = instance['citation']
    if len(m.split(';')) > 2:
        return 1
    return -1

cit_len_rule = WeakRule(
    exec_module=length_citation, 
    label_maps={0:[2], 1:[0,1]})
\end{minted}
\begin{minted}[framesep=1mm,
baselinestretch=0.4,
fontsize=\small
]{python}
def df3(instance):
    def result_related(inst):
        patterns = ['equivocal result', 
                    'similar result', 
                    'same result', 
                    'different result', 
                    'expected result']
        for pattern in patterns:
            if pattern in inst:
                return 1
        return -1
    return result_related(instance['preprocessed_string'].lower())

res_rule = WeakRule(
    exec_module=df3, 
    label_maps={0:[1], 1:[0,2]})
\end{minted}
\begin{minted}[framesep=1mm,
baselinestretch=0.4,
fontsize=\small
]{python}
re_patt_0 = 'using|measuring|used|the method|' +
'data|state-of-art|calculated|applied|' +
'according to|approach'
simple_re_rule_1 = BinaryRERules(
    re_pattern=re_patt_0, 
    preproc=lambda inst:inst['string'], 
    label_maps={0:[0,2], 1:[1]}, unipolar=False)
\end{minted}
\begin{minted}[framesep=1mm,
baselinestretch=0.4,
fontsize=\small
]{python}

def df4(instance):
    def comparison(inst):
        patterns = [
            'in line with', 
            'discordant with', 
            'consistent with', 
            'keeping with', 
            'accordance with', 
            'agreement with', 
            'similar with',
            'compared to', 
            'contrast to', 
            'contrary to', 
            'comparable to', 
            'contradict to', 
            'affirmed by', 
            'supported by', 
            'in support of', 
        ]
        for pattern in patterns:
            if pattern in inst:
                return 1
        return -1
    return comparison(instance['preprocessed_string'].lower())

resultp_rule = WeakRule(
    exec_module=df4, 
    label_maps={0:[0,1], 1:[2]})
\end{minted}
\begin{minted}[framesep=1mm,
baselinestretch=0.4,
fontsize=\small
]{python}

def df5(instance):    
    def match(inst):
        patterns = [
            'has been', 
            'in order to', 
            'considered to', 
            'initially', 
            'even if',
            'have shown', 
            'has shown'
        ]
        for pattern in patterns:
            if pattern in inst:
                return 1
        return -1
    return match(instance['preprocessed_string'].lower())

add_rule = WeakRule(
    exec_module=df5, 
    label_maps={0:[2], 1:[0,1]})

\end{minted}
\begin{minted}[framesep=1mm,
baselinestretch=0.4,
fontsize=\small
]{python}
re_patt_1 = 'our result|this is in keeping|' +
'with previous|this study differ|' +
'this is in agreement with|this conclusion|' + 
'this finding|' +
'similar result.*(found|observed|obtained)'
re_tb = BinaryRERules(
    re_pattern=re_patt_1, 
    preproc=lambda inst:inst['string'], 
    label_maps={0:[0,1], 1:[2]}, unipolar=True)
    \end{minted}
\begin{minted}[framesep=1mm,
baselinestretch=0.4,
fontsize=\small
]{python}
re_patt_2 = 'we (employ|utilize)|iap-as|' +
'metaanalyses|temporal transition|' + 
'was estimated|quantitative (analyses|analysis)' +
'|this procedure|implementation|sequence analysis' +
'|regularization method|were analysed|we adopt|' +
'bayesian|were sampled|quantitative method|fracture|' + 
simulating|this design|algorithm|developed|' + 
'model performance|(was|were) evaluated|as control' +
'|scheme|control management|(was|were|is|are) measured' +
'|the rats|the pigs|we appl'
re_tm = BinaryRERules(
    re_pattern=re_patt_2, 
    preproc=lambda inst:inst['string'], 
    label_maps={0:[0,2], 1:[1]}, unipolar=True)
\end{minted}

%% file: figs/plf_sep_trec.tex
\begin{minted}
    [
framesep=1mm,
baselinestretch=0.4,
fontsize=\small
]{python}

def first_word(instance):
    st = instance['string']
    word = st.split()[0].lower()
    if word == "who":
        return 0
    elif word == "where":
        return 1
    elif word == "when":
        return 2
    elif word == "why":
        return 3
    elif word == "how":
        return 4
    elif word == "name":
        return 5
    else:
        return -1
first_word_rule = WeakRule(
    exec_module=first_word, 
    label_maps={
        0: [3],
        1: [4],
        2: [5],
        3: [1],
        4: [1, 5],
        5: [2, 3],
        6: [0] 
    })

\end{minted}
\begin{minted}[framesep=1mm,
baselinestretch=0.4,
fontsize=\small
]{python}
def called(instance):
    st = instance['string'].lower().split()
    if "called" in st:
        return 1
    
    return -1
r1 = WeakRule(
    exec_module=called, 
    label_maps={
        0: [0, 1, 5],
        1: [2, 3, 4]
    })

\end{minted}
\begin{minted}[framesep=1mm,
baselinestretch=0.4,
fontsize=\small
]{python}
def mean(instance):
    st = instance['string'].lower().split()
    if "mean" in st or "meaning" in st:
        return 1
    return -1
r2 = WeakRule(
    exec_module=mean, 
    label_maps={
    0: [2, 3, 4, 5],
    1: [0, 1]
    })

\end{minted}
\begin{minted}[framesep=1mm,
baselinestretch=0.4,
fontsize=\small
]{python}
def abbre(instance):
    st = instance['string'].lower()
    if "stand for" in st or "abbreviat" in st:
        return 1
    
    return -1
r3 = WeakRule(
    exec_module=abbre, 
    label_maps={
        0: [1, 2, 3, 4, 5],
        1: [0]
    })

\end{minted}
\begin{minted}[framesep=1mm,
baselinestretch=0.4,
fontsize=\small
]{python}
def desc(instance):
    st = instance['string'].lower()
    tokens = st.split()
    if "definition" in tokens or \
            "come from" in st or \
            "origin" in tokens:
        return 1
    return -1
r4 = WeakRule(
    exec_module=desc, 
    label_maps={
        0: [0, 2, 3, 4, 5],
        1: [1]
    })
\end{minted}
\begin{minted}[framesep=1mm,
baselinestretch=0.4,
fontsize=\small
]{python}

def enty(instance):
    subject = get_subject(instance['string'])
    if subject in ("animal", "body", 
                   "color", "creative",
                   "currency", "disease", 
                   "event", "food", "instrument",
                   "language", "letter", "plant", 
                   "product", "religion",
                   "sport", "substance", "symbol", 
                   "technique", "term",
                   "vehicle", "word"):
        return 1
    return -1
r5 = WeakRule(
    exec_module=enty, 
    label_maps={
        0: [0, 1, 3, 4, 5],
        1: [2]
    })
\end{minted}
\begin{minted}[framesep=1mm,
baselinestretch=0.4,
fontsize=\small
]{python}

def loc(instance):
    subject = get_subject(instance['string'])
    if subject in ("city", "country", "mountain", "state", "capital"):
        return 1
    
    return -1
r7 = WeakRule(
    exec_module=loc, 
    label_maps={
        0: [0, 1, 2, 3, 5],
        1: [4]
    })

\end{minted}
\begin{minted}[framesep=1mm,
baselinestretch=0.4,
fontsize=\small
]{python}
def num(instance):
    st = instance['string'].lower()
    if "what year" in st:
        return 1
    
    return -1
r8 = WeakRule(
    exec_module=num, 
    label_maps={
        0: [0, 1, 2, 3, 4],
        1: [5]
    })
\end{minted}
\begin{minted}[framesep=1mm,
baselinestretch=0.4,
fontsize=\small
]{python}

def num1(instance):
    st = instance['string'].lower()
    if "how many" in st or 'how much' in st or 'how old' in st:
        return 1
    
    return -1
r9 = WeakRule(
    exec_module=num1, 
    label_maps={
        0: [0, 1, 2, 3, 4],
        1: [5]
    })

\end{minted}
\begin{minted}[framesep=1mm,
baselinestretch=0.4,
fontsize=\small
]{python}
whatmean = BinaryRERules(
    name='what_*_mean',
    re_pattern='what.*mean', 
    preproc=lambda inst:inst['string'].lower(), 
    label_maps={0:[0, 2, 3, 4,5], 1:[1]}, unipolar=True)

\end{minted}
\begin{minted}[framesep=1mm,
baselinestretch=0.4,
fontsize=\small
]{python}
desc_r1 = BinaryRERules(
    name='what_*_',
    re_pattern='what.*use of|what.*origin of|why do', 
    preproc=lambda inst:inst['string'].lower(), 
    label_maps={0:[0, 2, 3, 4,5], 1:[1]}, unipolar=True)

\end{minted}
\begin{minted}[framesep=1mm,
baselinestretch=0.4,
fontsize=\small
]{python}
numpattern1 = BinaryRERules(
    name='num_patt',
    re_pattern='how far|what.*birthday|how long|how deep|'+
    'when did|when was|how tall|what month|population|toll'+
    '|how big|how long|what year', 
    preproc=lambda inst:inst['string'].lower(), 
    label_maps={0:[0,1, 2, 3, 4], 1:[5]}, unipolar=True)
\end{minted}
\begin{minted}[framesep=1mm,
baselinestretch=0.4,
fontsize=\small
]{python}

descpatt_1 = BinaryRERules(
    name='num_patt',
    re_pattern='what is the origin|what is the history'+
    '|what.*mean|how do you buy|what is the difference'+
    '|how can I|how do I|what effect', 
    preproc=lambda inst:inst['string'].lower(),      
    label_maps={0:[0, 2, 3, 4,5], 1:[1]}, 
    unipolar=True)
\end{minted}
\begin{minted}[framesep=1mm,
baselinestretch=0.4,
fontsize=\small
]{python}
 
descpatt_2 = BinaryRERules(
    name='num_patt',
    re_pattern='how.*tell|how d.*affect|how do.*work'+
    '|how do you fix|how do you get|how do you find'+
    '|how do I find|how.*made', 
    preproc=lambda inst:inst['string'].lower(), 
    label_maps={0:[0, 2, 3, 4,5], 1:[1]}, unipolar=True)

\end{minted}
\begin{minted}[framesep=1mm,
baselinestretch=0.4,
fontsize=\small
]{python}
newhow = BinaryRERules(
    name='num_patt',
    re_pattern='how do|how was|how are|how is'+
    '|how was|how could|how can', 
    preproc=lambda inst:inst['string'].lower(), 
    label_maps={0:[0, 2, 3, 4,5], 1:[1]}, unipolar=True)

\end{minted}
\begin{minted}[framesep=1mm,
baselinestretch=0.4,
fontsize=\small
]{python}
wsf = BinaryRERules(
    name='what_*_sf',
    re_pattern='what.*stand for', 
    preproc=lambda inst:inst['string'].lower(), 
    label_maps={0:[1, 2, 3, 4,5], 1:[0]}, unipolar=True)
\end{minted}

%% file: figs/plfs_sep_ag.tex
\begin{minted}
    [
framesep=1mm,
baselinestretch=0.4,
fontsize=\small
]{python}

def df1(instance):
    entity_doc = instance['sen_ner']
    if 'EVENT' in  entity_doc and 'GPE' in entity_doc:
        return 1
    return -1
gpe_event_title = WeakRule(
    name='gpe+event_title', 
    exec_module=df1, 
    label_maps={0:[0,2,3], 1:[1]})

\end{minted}
\begin{minted}[framesep=1mm,
baselinestretch=0.4,
fontsize=\small
]{python}
def df2(instance):
    entity_doc = instance['title_ner']
    if 'LOC' in entity_doc:
        return 1
    return -1
loc_title_rule = WeakRule(name='loc_title', 
    exec_module=df2, l
    abel_maps={0:[1,2,3], 1:[0]})

\end{minted}
\begin{minted}[framesep=1mm,
baselinestretch=0.4,
fontsize=\small
]{python}
sp0 = BinaryRERules(
    name='sports_names',
    re_pattern='kelvim escobar|red sox|'+
    'formula one|grand prix|svetlana kuznetsova'+
    '|billy wagner| kobe | beckham|johnny damon'+
    '|robin ventura|olivier panis', 
    preproc=lambda inst:inst['sen_lemma'].lower(), 
    label_maps={0:[0,2,3], 1:[1]}, unipolar=True)
\end{minted}
\begin{minted}[framesep=1mm,
baselinestretch=0.4,
fontsize=\small
]{python}

def bexclusive_lemmas(instance):
    business_keywords = {'profit', 'bankrupt', 'yen', 'financial'}
    doc = instance['sen_lemma'].lower().split()
    for word in doc:
        for keyword in business_keywords:
            if word.startswith(keyword):
                return 1
    return -1
bexclusive_lemmas_rule = WeakRule(
    name='bus_lemma', 
    exec_module=bexclusive_lemmas, 
    label_maps={0:[0,1,3], 1:[2]})

\end{minted}
\begin{minted}[framesep=1mm,
baselinestretch=0.4,
fontsize=\small
]{python}
tech_patt = 'space.com|space station|'+
    'network authentication|(python|java|matlab|c) developer|'+
    'application|virus|browser hijack|search engine|internet-based|'+
    'windows update|smart phone|source code|mangement software'+
    '|software.*develop|internet connection|interactive gam'+
    '|game console|transfer datum|internet security|g network'+
    '|internet company|storage capacity|music player|microsystem'+
    '|comsumer electronic|operat.*system|wireness network|motherboard'+
    '|spacecraft|malicious program|video game'
techr = BinaryRERules(name='tech_terms',
    re_pattern=tech_patt,                      
    preproc=lambda inst:inst['sen_lemma'].lower(),                          
    label_maps={0:[0,1,2], 1:[3]}, unipolar=True)

\end{minted}

\begin{minted}[framesep=1mm,
baselinestretch=0.4,
fontsize=\small
]{python}

def exclusive_lemmas(instance):
    
    world_pol_keywords = {'mideast', 'iraq', 
                          'baghdad', 'pakistan', 
                          'afghan', 'kurd', 'arab',
                          'egypt', 'iran', 'turkey',
                          'syria', 'bahrain', 
                          'israel', 'jordan', 
                          'kuwait', 'lebanon', 
                          'oman', 'palestine', 
                          'qatar', 'saudi', 'uae', 
                          'yemen' ,'chechnya'}
    world_pol_keywords |= {'al-qaeda', 'taliban'}
    world_pol_keywords |= {'hostage', 'abduct', 
                            'hijack'}
    world_pol_keywords |= {'invasion', ' coup ', 
                            'curfew', 'army', 'troop', 
                            'peace', 'militant', 'missile'}
    world_pol_keywords |= {'murder', 'death'}
    sports_keywords = {'baseball', 'football', 'soccer',
                        'hockey', 'basketball', 
                            'tennis', 'golf'}
    sports_keywords |= {'stadium', 'arena'}
    sports_keywords |= {'season', 'playoff', 'tournament'}
    sports_keywords |= {'mlb', 'nfl', 'nba', 'mls', 
                        'nhl', 'ncaa', 'league', 'racing'}
    sports_keywords |= {'premiership'}
    sports_keywords |= {'quarterback', 'centerback', 
                        'fullback', 'pitcher'}
    business_keywords = {'profit', 'bankrupt', 'financial'}
    bt_keywords = {'robot', 'robotic'}
    bt_keywords |= {'web', 'internet'}
    bt_keywords |= {'linux'}
    bt_keywords |= {'stem-cell', 'biotechnology'}
    bt_keywords |= {'xbox', 'playstation'}
    bt_keywords |= {'microsoft'}
    bt_keywords |= {'space', 'nasa'}
    bt_keywords |= {'adobe', 'ipod', 'apple', 'xerox', 'ibm'}
    
    doc = instance['sen_lemma'].lower().split()
    
    for word in doc:
        for keyword in bt_keywords:
            if word.startswith(keyword):
                return 3
            
    for word in doc:
        for keyword in world_pol_keywords:
            if word.startswith(keyword):
                return 0

    for word in doc:
        for keyword in sports_keywords:
            if word.startswith(keyword):
                return 1
            
    for word in doc:
        for keyword in business_keywords:
            if word.startswith(keyword):
                return 2
    return -1
exclusive_lemmas_rule = WeakRule(
    exec_module=exclusive_lemmas, 
    label_maps={0:[0], 1:[1], 2:[2], 3:[2,3]})
\end{minted}
\begin{minted}[framesep=1mm,
baselinestretch=0.4,
fontsize=\small
]{python}
def df3(instance):
    entity_doc = instance['title_ner']
    if 'PERSON' in  entity_doc and 'GPE' in entity_doc:
        return 1
    return -1
person_gpe_title_rule = WeakRule(
    name='p+gpe_title', 
    exec_module=df3, 
    label_maps={0:[2, 3], 1:[0, 1]})
\end{minted}
\begin{minted}[framesep=1mm,
baselinestretch=0.4,
fontsize=\small
]{python}

def df4(instance):
    entity_doc = instance['sen_ner']
    if 'PERSON' in  entity_doc and 'EVENT' in entity_doc:
        return 1
    return -1
person_event_sen_rule = WeakRule(
    name='person+event_sen', 
    exec_module=df4, 
    label_maps={0:[2,3], 1:[0,1]})

    \end{minted}
\begin{minted}[framesep=1mm,
baselinestretch=0.4,
fontsize=\small
]{python}
def df5(instance):
    entity_doc = instance['title_ner']
    if  'PRODUCT' in entity_doc:
        return 1
    return -1
sen_prod_t_rule = WeakRule(
    exec_module=df5, 
    label_maps={0:[0,1], 1:[2,3]})

\end{minted}
\begin{minted}[framesep=1mm,
baselinestretch=0.4,
fontsize=\small
]{python}
dpw2 = BinaryRERules(
    name='dpw2',
    re_pattern='minister|chairman', 
    preproc=lambda inst:inst['sen_lemma'].lower(), 
    label_maps={0:[1,3], 1:[0,2]}, unipolar=True)

\end{minted}
\begin{minted}[framesep=1mm,
baselinestretch=0.4,
fontsize=\small
]{python}
intnr = BinaryRERules(
    name='tech',
    re_pattern='internet', 
    preproc=lambda inst:inst['sen_lemma'].lower(), 
    label_maps={0:[1,2], 1:[0,3]}, unipolar=True)
\end{minted}